\documentclass{article}

\usepackage{titlesec}
\usepackage{graphicx}
\usepackage{arxiv}
\usepackage[utf8]{inputenc}
\usepackage{setspace}
\usepackage{geometry}
\usepackage{indentfirst}
\usepackage{amsmath,amssymb,amsfonts}
\usepackage{xcolor}
\definecolor{verylightgray}{gray}{0.9}

\usepackage{booktabs,tabularx,makecell}
\usepackage{booktabs}
\usepackage{multirow}
\newcolumntype{Y}{>{\raggedright\arraybackslash}X}
\newcolumntype{L}{>{\raggedright\arraybackslash}X}
\usepackage{array}
\usepackage{float}
\usepackage{algorithm}
\usepackage{algorithmic}

\usepackage{natbib}
\usepackage[T1]{fontenc}
\usepackage{hyperref}
\usepackage{url}
\usepackage{nicefrac}
\usepackage{microtype}
\usepackage{doi}
\usepackage{cleveref}
\newcommand{\method}{PEARL}
\newcommand{\R}{\mathbb{R}}
\newcommand{\E}{\mathbb{E}}
\newcommand{\Loss}{\mathcal{L}}
\newcommand{\Data}{\mathcal{D}}

\title{PEARL: Prototype-Enhanced Alignment for Label-Efficient Representation Learning with Deployment-Driven Insights from Digital Governance Communication Systems}
\date{}

\author{
	\href{https://orcid.org/0000-0002-0883-4574}{\includegraphics[scale=0.06]{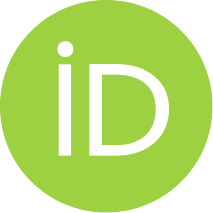}\hspace{1mm}Ruiyu Zhang}\\
	Department of Politics and Public Administration\\
    The University of Hong Kong\\
	ruiyuzh@connect.hku.hk\\
	\And
    	\href{https://orcid.org/0000-0002-0275-117X}{\includegraphics[scale=0.06]{orcid.pdf}\hspace{1mm}Lin Nie}\\
	Department of Applied Social Sciences\\
    The Hong Kong Polytechnic University\\
    lin-apss.nie@polyu.edu.hk\\
    \And
	\href{https://orcid.org/0000-0002-2303-0996}{\includegraphics[scale=0.06]{orcid.pdf}\hspace{1mm}Wai-Fung Lam}\\
	Department of Politics and Public Administration\\
    The University of Hong Kong\\
    dwflam@hku.hk\\
    \And
	\href{https://orcid.org/0009-0005-0284-1289}{\includegraphics[scale=0.06]{orcid.pdf}\hspace{1mm}Qihao Wang}\\
	School of Physical Science and Technology\\
    Lanzhou University\\
    wangqh2021@lzu.edu.cn\\
    \And
	\href{https://orcid.org/0009-0005-7399-109X}{\includegraphics[scale=0.06]{orcid.pdf}\hspace{1mm}Xin Zhao}\\
	Department of Applied Social Sciences\\
    The Hong Kong Polytechnic University\\
    xinnn.zhao@connect.polyu.hk
}

\renewcommand{\headeright}{ }
\renewcommand{\undertitle}{ }
\renewcommand{\shorttitle}{PEARL: Prototype-Enhanced Alignment for Label-Efficient Representation Learning}

\hypersetup{
    pdftitle={PEARL: Prototype-Enhanced Alignment for Label-Efficient Representation Learning},
    pdfauthor={Ruiyu Zhang, Lin Nie, Wai-Fung Lam, Qihao Wang, Xin Zhao},
    pdfkeywords={Representation Geometry, Embedding Post-processing, Prototype Alignment, Label Efficiency, Retrieval, Digital Governance},
}

\begin{document}
\newgeometry{
    a4paper,
    left=0.9in,
    right=0.9in,
    top=0.9in,
    bottom=0.8in,
}
\maketitle
\footnotetext[1]{The PEARL Python package can be installed via: \colorbox{verylightgray}{\texttt{pip install pearl-H}}. We provide PEARL as an off-the-shelf, easy-to-use package so that researchers can directly apply it in their work. For documentation and tutorials, visit: \url{https://pypi.org/project/pearl-H/}.}

\begin{abstract}
\noindent
In many deployed systems, new text inputs are handled by retrieving similar past cases—for example, routing and responding to citizen messages in digital governance platforms. When these systems fail, the issue is often not the language model itself, but that the nearest neighbors in the embedding space correspond to the wrong cases. Modern machine learning systems increasingly rely on fixed, high-dimensional embeddings produced by large pretrained models and sentence encoders. In real-world deployments, labels are scarce, domains shift over time, and retraining the base encoder is expensive or infeasible. As a result, downstream performance depends heavily on embedding geometry. Yet raw embeddings are often poorly aligned with the local neighborhood structure required by nearest-neighbor retrieval, similarity search, and lightweight classifiers that operate directly on embeddings. We propose \textbf{\method} (Prototype-Enhanced Aligned Representation Learning), a label-efficient approach that uses limited supervision to softly align embeddings toward class prototypes. The method reshapes local neighborhood geometry while preserving dimensionality and avoiding aggressive projection or collapse. Its aim is to bridge the gap between purely unsupervised post-processing, which offers limited and inconsistent gains, and fully supervised projections that require substantial labeled data. We evaluate \method\ under controlled label regimes ranging from extreme label scarcity to higher-label settings. In the label-scarce condition, \method\ substantially improves local neighborhood quality, yielding 25.7\% gains over raw embeddings and over 21.1\% gains relative to strong unsupervised post-processing, precisely in the regime where similarity-based systems are most brittle. As label availability increases, fully supervised projections can achieve stronger performance on some tasks, while \method\ remains a robust and practical preprocessing step that delivers reliable improvements with minimal supervision.\\
\vspace{0.5em}

\end{abstract}

\restoregeometry

\section{Introduction}
\label{sec:introduction}

Modern systems frequently treat pretrained embeddings as \emph{fixed features} for downstream tasks such as nearest-neighbor retrieval, similarity search, lightweight classifiers (kNN/centroid/linear probes), and decision support \citep{peters2019tune, reimers2019sbert, cer2018use}. In many settings the encoder is costly to retrain or cannot be modified, so downstream performance depends heavily on the geometry of the embedding space \citep{ethayarajh2019contextual, li2020sentence}. Recent work has also developed general-purpose sentence embedding models via instruction finetuning or weakly supervised contrastive pretraining, aiming to produce more stable and transferable embedding spaces without task-specific fine-tuning \citep{su2023instructor, ni2022sentencet5, wang2022e5}. Yet regardless of how embeddings are produced, their geometry matters most when the downstream method is similarity-driven. For example, kNN-style prediction \citep{cover1967nearest} and memory-augmented inference \citep{khandelwal2020generalization} effectively \emph{delegate} a large part of the computation to the embedding space: the model retrieves neighbors and then aggregates their labels or responses. In such pipelines, small distortions in neighborhood structure can lead to systematically wrong retrieval and brittle behavior in rare or newly emerging categories.

The practical reality of many deployments reinforces this concern. Embeddings are often precomputed and cached (or produced by a shared foundation model service) and then consumed by multiple downstream applications, making end-to-end fine-tuning expensive and organizationally difficult. A lightweight, label-efficient post-processing step that improves neighborhood structure without changing the base encoder can therefore have outsized impact.

A central challenge is that raw embeddings are not necessarily optimized for downstream neighbor quality. Simple post-processing (centering, L2 normalization, PCA, common component removal) can help in some cases but often yields limited or inconsistent improvements \citep{arora2017sif, stankevivcius2024extracting}. At the other extreme, supervised projection methods (e.g., LDA) can be strong when labels are abundant, but may become unstable or data-hungry in low-label regimes \citep{snell2017prototypical, vogelstein2021supervised}. This creates a gap between unsupervised geometric fixes and fully supervised discriminative projections. Importantly, the challenge is not merely ``separating classes'' in a global sense. In many systems, what matters is the \emph{local} structure: whether nearest neighbors share labels, whether semantically similar items appear early in the ranked list, and whether the embedding space supports robust case retrieval under distribution shift. In digital governance, for example, citizens may use similar phrasing across different complaint categories, creating shared stylistic factors that dominate cosine similarity even when the underlying intent differs. A useful method must therefore do three things simultaneously: (i) leverage the limited label information that is available, (ii) target local neighborhoods and early retrieval precision, and (iii) avoid aggressively collapsing the representation, which would harm reuse across tasks and reduce interpretability for human-in-the-loop workflows. This raises a key question: can we reshape representation geometry in a label-efficient way that improves nearest-neighbor quality and early retrieval precision, works reliably when labels are scarce, and remains compatible with downstream supervised methods?

We introduce {\method} (Prototype-Enhanced Aligned Representation Learning), a Prototype-Enhanced Alignment method for label-efficient geometry shaping. \method\ uses limited label information to align embeddings toward class prototypes while preserving dimensionality and avoiding hard projection. We evaluate \method\ against standard post-processing baselines and supervised projections under controlled label availability, and demonstrate its practical relevance in digital governance communication systems. Concretely, this paper contributes a label-efficient geometry shaping objective that aligns embeddings toward class prototypes while controlling collapse through reconstruction and regularization; a systematic evaluation protocol that separates neighborhood quality (Purity@K, separation) from retrieval ranking metrics (Hit@K, MRR@K) across label regimes; and evidence from a high-stakes deployment context (digital governance communication) that motivates why early retrieval precision matters and why low-label operation is the norm rather than the exception.

\section{Motivation and Application Context}
\label{sec:context}

Governments increasingly rely on digital platforms to communicate with citizens (e.g., online complaint systems, consultation portals, public service feedback channels). These systems generate large volumes of unstructured text, where embedding-based retrieval and classification are common for routing messages, detecting response types, and matching queries to relevant policies and services. From a system perspective, many governance workflows are naturally ``retrieval-first'': a message arrives, the system retrieves similar historical cases (or frequently asked questions), and then suggests (i) a candidate category, (ii) relevant policy articles, and (iii) a recommended handling unit. This retrieval step is valuable even when the final decision is made by humans, because it supports transparency and auditability---civil servants can inspect similar cases and assess whether the model's recommendation is reasonable \citep{aamodt1994cbr, das2021case}. More broadly, recent work suggests that AI can reshape citizen--government interactions in ways that matter for service quality and institutional practice \citep{zhang2025enhancing, dunleavy2006deg}, reinforcing the importance of robust, label-efficient representation methods in governance settings where systems must support reliable retrieval and triage under practical constraints.

At the same time, public administration research highlights both opportunities and governance challenges of deploying AI in the public sector, including implementation constraints, accountability concerns, and the role of discretion in bureaucratic work \citep{sun2019ai_public, wirtz2019ai_public, bullock2019ai_discretion, yeung2018algorithmic, danaher2017algorithmic}. These perspectives motivate methods that improve system behavior under label scarcity and distribution shift, rather than assuming abundant supervision and static environments.

However, this reliance on retrieval makes the embedding space a critical piece of infrastructure. Two messages can be lexically similar but belong to different procedural categories (e.g., ``request for information'' vs.\ ``complaint''), and the system must separate these patterns without extensive retraining. When the embedding space overemphasizes shared style and underemphasizes category-discriminative content, the retrieved set can be misleading even if the top-level classifier appears to perform adequately on average. Digital governance communication also captures a realistic mix of phenomena that stress-test representation geometry: domain shift due to evolving policies, long-tailed intents, class imbalance, and heterogeneous language styles (formal submissions, informal complaints, copied templates, or short follow-ups). These factors motivate a method that directly targets neighborhood structure under limited labels.

These systems also operate under practical constraints that shape what methods are feasible. Labels are costly and slow to produce (requiring trained civil servants or policy experts), labeling standards evolve, and distributions are non-stationary as new topics emerge and language shifts. In practice, systems often need to reuse existing embeddings and improve performance with minimal retraining. Beyond annotation cost, governance settings introduce operational constraints that further shape method design: data access can be sensitive (logs may contain personal information, which restricts experimentation and limits the feasibility of frequent end-to-end retraining), and workflows are often multi-stage and involve multiple agencies. As a result, embeddings extracted from text may be shared across applications (triage, routing, monitoring, summarization), and aggressive task-specific fine-tuning can create regressions in other components. Moreover, governance systems must accommodate continuous change---a small amount of new labeling may become available when a new topic emerges or when a campaign is launched (e.g., seasonal public service requests). A label-efficient method should make effective use of such incremental supervision: rather than requiring a full retraining cycle, it should be able to improve local retrieval behavior quickly. These issues can be understood as a form of dataset shift, where the joint distribution of text and labels changes over time as policies, topics, and user language evolve \citep{quinonero2009datasetshift}. Methods that rely on abundant labeled data or frequent end-to-end retraining are therefore less practical than label-efficient approaches that can be updated incrementally. Relatedly, governance and political text analysis often requires semantic \emph{consistency} over time (rather than assuming semantic change), which makes stability of contextual representations an important practical consideration \citep{zhang2024achieving}.

Importantly, these constraints are not unique to governance---the same patterns appear in many industry settings, including customer support, enterprise knowledge bases, compliance monitoring, and healthcare triage. Digital governance is a representative high-stakes application, not a special case. In customer support and enterprise search, for instance, organizations often deploy a shared embedding service and then build multiple downstream tools on top: semantic search, ticket routing, and recommendation. Labels for new products, new policy rules, or new user segments appear gradually. Similarly, compliance and monitoring systems face evolving terminology and shifting distributions, with limited labeled examples for new risk categories. Across these domains, the key failure mode is consistent: when labels are scarce, the system increasingly falls back on similarity-based reasoning, and therefore the quality of \emph{local neighborhoods} becomes the bottleneck. This is precisely the regime targeted by \method.

\section{Related Work}
\label{sec:related}

We position \method\ relative to four strands of literature: (i) unsupervised embedding post-processing, (ii) prototype and centroid-based methods, (iii) supervised projections and metric learning, and (iv) retrieval-oriented systems that depend on neighborhood structure.

One common approach is embedding post-processing via centering and normalization, PCA-based dimensionality reduction, and whitening variants \citep{su2021whitening}. These methods are largely unsupervised: they reshape global statistics but do not directly target semantic neighborhood structure under limited labels. Unsupervised post-processing is attractive because it is simple and model-agnostic---it can be applied to any embedding matrix without labels. In practice, it is often used to reduce anisotropy, stabilize cosine similarity, and accelerate retrieval by improving variance structure \citep{ethayarajh2019contextual, li2020sentence}. Whitening methods, for example, attempt to re-scale directions of high variance so that similarity is less dominated by a small number of common components \citep{su2021whitening}, while other widely used techniques remove dominant common components (``all-but-the-top'') or compute smoothed sentence representations to improve isotropy \citep{arora2017sif, mu2018allbuttop}. However, unsupervised operations cannot know which directions correspond to class-discriminative signals for a particular downstream task. In a low-label regime, it is common that the directions most responsible for retrieval errors are correlated with spurious factors (formatting, politeness templates, or shared bureaucratic phrasing). Correcting these errors benefits from even a small amount of supervision that reveals which neighborhoods should be pulled together or pushed apart.

A second line of work relies on prototypes or centroids, appearing in classic relevance feedback and centroid-based classification \citep{rocchio1971relevance, mensink2013ncm} as well as few-shot learning \citep{snell2017prototypical}. Such methods can be effective but are often tightly coupled to a specific classifier or risk collapsing representations when used as a global transformation. Prototypes provide a simple and interpretable representation of a class---the prototype summarizes what is common among labeled examples---and in few-shot settings, prototype-based inference can be remarkably effective because it reduces the problem to comparing a query against a small set of centroids \citep{snell2017prototypical}. In operational systems, this interpretability is also valuable: prototypes can be inspected, tracked over time, and audited for drift. At the same time, prototypes alone are not a complete solution. If we simply replace embeddings with their nearest prototype, we lose within-class variation and harm downstream retrieval and ranking. Conversely, if we only use prototypes as classifier weights, we do not explicitly improve neighborhood structure for the broader retrieval tasks that motivate this work. \method\ uses prototypes as \emph{anchors} for a transformation that preserves dimensionality and retains instance-level structure.

A third line of work uses supervised projections and metric learning. Supervised projections can yield strong performance with abundant labels, but they optimize separability in a way that may be unstable under label scarcity and are not designed as a general-purpose preprocessing step for cosine retrieval \citep{li2023deep, song2023comprehensive}. These projections are typically optimized to separate classes globally, often with objectives that emphasize between-class scatter relative to within-class scatter. When labels are plentiful, this can yield large gains for classification, but in low-label regimes, the estimated class statistics are noisy, and the projection may overfit to a small sample of labeled points, producing unstable behavior (especially for rare classes). In addition, supervised projections are usually applied as a fixed linear map, which may not capture the kinds of nonlinear distortions present in high-dimensional embeddings produced by pretrained models. This motivates a middle ground: use limited supervision to guide a lightweight transformation that targets neighborhood structure, without requiring the full labeled data needs of traditional supervised metric learning \citep{kulis2013metric, weinberger2009lmnn, schroff2015facenet}. More broadly, metric learning and supervised contrastive learning shape representations by pulling together points of the same class and pushing apart points of different classes \citep{kulis2013metric, khosla2020supcon, hadsell2006dimensionality, sohn2016npair}. PEARL shares the local-geometry motivation of these methods, but differs in two ways: (i) it uses prototypes as low-variance anchors when labels are scarce, and (ii) it explicitly preserves the original embedding dimensionality and information content for reuse across retrieval pipelines. Related disentanglement ideas also appear in deep generative modeling, where architectural structure can be used to align latent subspaces with interpretable factors and to isolate nuisance variation \citep{zhang2025structural}. While PEARL is not a generative model, the same design philosophy applies: separating task-relevant signal from confounding variation can be easier to achieve through explicit architectural factorization than through regularization alone.

Taken together, these strands motivate a middle ground between purely unsupervised fixes and fully supervised projections. \method\ sits in this gap: it uses minimal supervision to \emph{reshape neighborhoods} rather than to fully optimize a discriminative classifier. This positioning is deliberate. In the applications we care about, the same embeddings may be used across multiple downstream tasks, and improving retrieval neighborhoods is often more valuable than squeezing out a small classification gain on a particular benchmark split. \method\ is therefore evaluated with geometry and retrieval diagnostics in addition to (or instead of) a single aggregate classification score. This focus is also aligned with the broader trend of retrieval-augmented systems, where downstream behavior is mediated by a retrieval step over an embedding index \citep{guu2020retrieval, lewis2020retrieval, zhong2022training}. While much of that literature focuses on retrieval-augmented generation, the core issue is shared: if retrieval neighborhoods are poorly aligned with task labels, the retrieved evidence can be misleading even when the base encoder is strong. In neural information retrieval, dense retrievers (e.g., DPR) and late-interaction models (e.g., ColBERT) explicitly optimize retrieval effectiveness over embedding spaces \citep{karpukhin2020dpr, khattab2020colbert}, and evaluation suites such as BEIR emphasize robustness across heterogeneous retrieval tasks \citep{thakur2021beir}. PEARL complements these lines of work by targeting label-efficient neighborhood shaping when retraining or large-scale IR supervision is not feasible.

\section{Method: \method}
\label{sec:method}

\method\ is designed to (i) preserve embedding dimensionality, (ii) improve local neighborhood structure, (iii) remain compatible with cosine-based retrieval, and (iv) compose with downstream supervised methods. We further emphasize two pragmatic principles: first, \method\ should be \emph{small}, meaning the refinement model should be lightweight compared to the base encoder, enabling fast training when a modest amount of labeling becomes available; second, \method\ should be \emph{stable under scarcity}, so that when only a few labels are available, the method improves early retrieval precision without depending on fragile estimates of high-order class statistics.

	Intuitively, each class has a prototype representing its semantic center. Raw embeddings can be scattered around these prototypes with noisy geometry, so nearest neighbors may be label-inconsistent. \method\ uses limited labels to encourage a soft alignment toward class prototypes while avoiding hard projection that would collapse the global space. The key intuition is that, under limited labels, a prototype offers a low-variance estimate of ``where the class is'' in embedding space. If we can gently increase the similarity between each labeled example and its class prototype while reducing similarity to other prototypes, we can reshape neighborhoods to be more label-consistent. Crucially, this does not require learning a full classifier boundary; it only requires learning how to tilt the embedding vectors in a locally meaningful way. This idea connects to broader themes in representation learning: useful representations factorize sources of variation, so that task-relevant factors align with similarity. \method\ operationalizes this goal for the specific case of fixed embeddings and scarce labels, by using prototypes as a weak supervision signal about which factor (class identity) should be emphasized in the geometry.

	\begin{figure}[t]
	    \centering
	    \includegraphics[width=\linewidth]{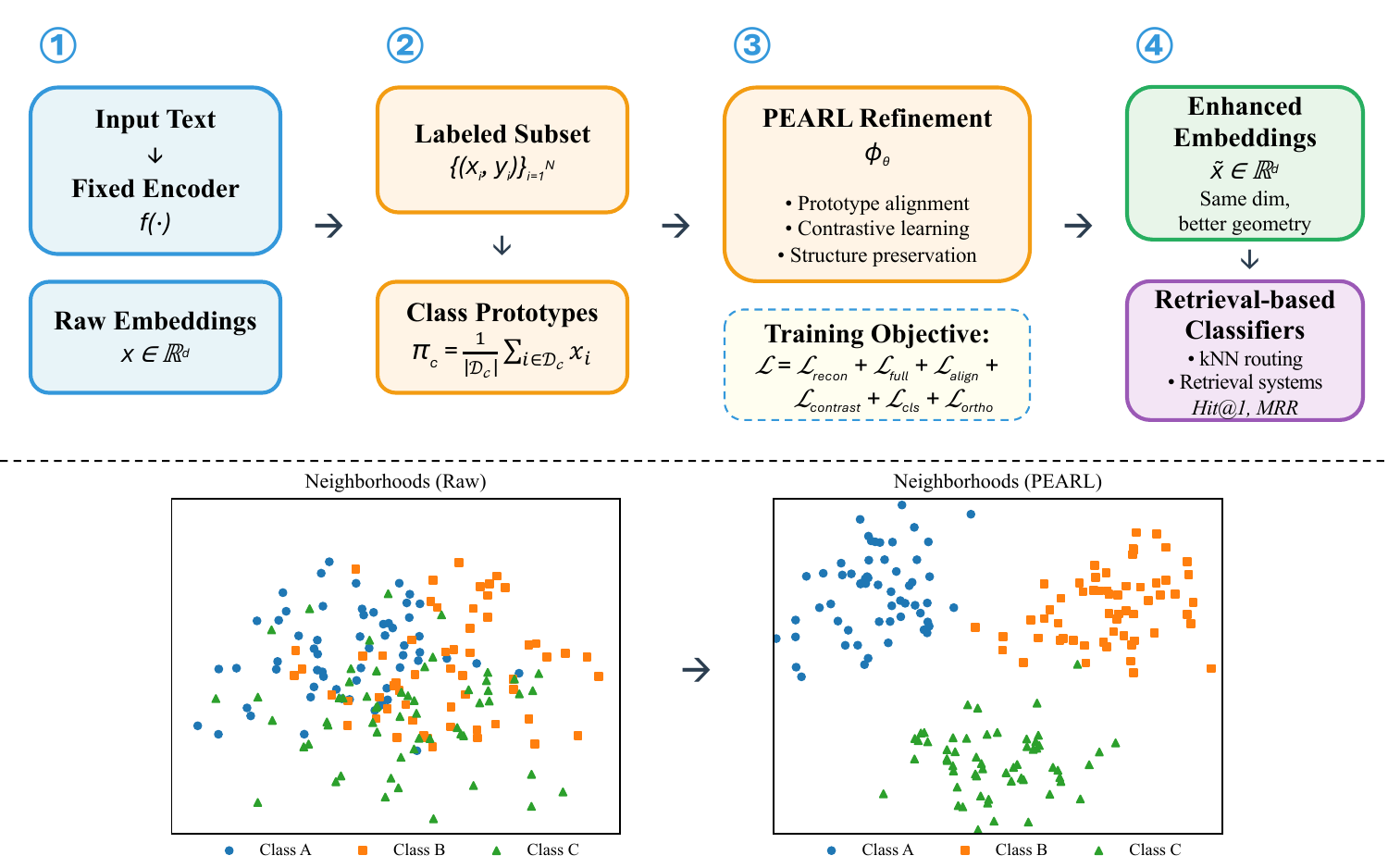}
	    \caption{Neighborhood geometry illustration. Raw embeddings can form overlapping neighborhoods, while \method\ reshapes local neighborhoods toward class prototypes to improve label-consistent retrieval under scarce supervision, without changing the embedding dimensionality.}
	    \label{fig:pearl_neighborhoods}
	\end{figure}
	
	Formally, let $X \in \R^{n \times d}$ be an embedding matrix and let $y$ denote a limited label set. For each class $c$, we compute a prototype as the mean of labeled embeddings in that class:
	\begin{equation}
	    \boldsymbol{\pi}_c = \frac{1}{|\Data_c|}\sum_{i \in \Data_c} \mathbf{x}_i,\quad
    \hat{\boldsymbol{\pi}}_c = \frac{\boldsymbol{\pi}_c}{\|\boldsymbol{\pi}_c\|_2}.
\end{equation}
We learn a refinement mapping $\phi_\theta: \R^d \rightarrow \R^d$ that preserves dimensionality and improves neighborhood structure. Training encourages three behaviors:
it increases similarity to the class prototype (prototype alignment), reduces similarity to other prototypes (prototype contrast), and preserves global structure via reconstruction and regularization to avoid collapse.
At inference time, \method\ outputs enhanced embeddings $\tilde{\mathbf{x}} = \phi_\theta(\mathbf{x})$ with the same dimensionality as the original embeddings and compatible with cosine similarity.

While \method\ can be implemented in different ways, one effective instantiation is a lightweight neural refinement model that extracts a class-discriminative ``signal'' component while separating residual variation. Concretely, we parameterize two encoders and a decoder:
\begin{align}
    \mathbf{z}_s &= E_s(\mathbf{x}) \in \R^{d_s}, \\
    \mathbf{z}_r &= E_r(\mathbf{x}) \in \R^{d_r}, \\
    \tilde{\mathbf{x}} &= D_s(\mathbf{z}_s) \in \R^d,
\end{align}
and optionally a full decoder $\hat{\mathbf{x}} = D_{\mathrm{full}}([\mathbf{z}_s;\mathbf{z}_r])$ to encourage information preservation. The enhanced embedding used downstream is $\tilde{\mathbf{x}}$, which preserves the original dimensionality and can be L2-normalized for cosine retrieval.

In the implementation, we additionally use a centroid prediction head $\mathrm{Proj}(\mathbf{z}_s)\in\R^d$ (L2-normalized) for prototype alignment and contrast, and a lightweight classifier $g(\mathbf{z}_s)\in\R^C$ to stabilize training and support early stopping. We train $\phi_\theta$ using a weighted combination of objectives that reflect the behaviors above:
\begin{align}
    \Loss_{\mathrm{recon}} &= \|D_s(\mathbf{z}_s) - \mathbf{x}\|_2^2, \\
    \Loss_{\mathrm{full}} &= \|D_{\mathrm{full}}([\mathbf{z}_s;\mathbf{z}_r]) - \mathbf{x}\|_2^2, \\
    \Loss_{\mathrm{align}} &= 1 - \cos(\mathrm{Proj}(\mathbf{z}_s), \hat{\boldsymbol{\pi}}_{y}), \\
    \Loss_{\mathrm{contrast}} &= -\log \frac{\exp(\cos(\mathrm{Proj}(\mathbf{z}_s), \hat{\boldsymbol{\pi}}_{y})/\tau)}{\sum_{c=1}^{C} \exp(\cos(\mathrm{Proj}(\mathbf{z}_s), \hat{\boldsymbol{\pi}}_{c})/\tau)}, \\
    \Loss_{\mathrm{cls}} &= \mathrm{CE}(g(\mathbf{z}_s), y), \\
    \Loss_{\mathrm{ortho}} &= \mathrm{mean}\!\left(\left|\bar{\mathbf{z}}_s^\top \bar{\mathbf{z}}_r\right|\right),
\end{align}
where $\mathrm{Proj}$ maps the signal to $\R^d$ for comparison with prototypes, $\tau$ is a temperature (set to $\tau=0.1$ in the our setting), and $\bar{\cdot}$ denotes row-wise normalization in a mini-batch. The total loss is a weighted sum of these terms; in the released package defaults, the full-reconstruction term is down-weighted (weight $0.5$) and the remaining weights are configurable.

Preserving the original dimensionality makes \method\ easy to integrate into existing systems: it can be applied as a drop-in replacement in any component that already consumes embeddings (indexing, clustering, kNN, or downstream models), and this also enables composition with supervised methods in higher-label regimes by applying them on top of \method\ outputs. Computationally, PEARL is lightweight relative to the base encoder: prototype computation is a single pass over labeled embeddings, training the refinement mapping scales linearly with the number of labeled points and the dimensionality of the embeddings, and can typically be done with small batch sizes because the model operates on vectors rather than token sequences. At inference time, the transformation is a single forward pass per embedding, which is inexpensive compared to generating the embedding itself. This makes it feasible to (i) transform embeddings on-the-fly for incoming messages, or (ii) periodically batch-transform the historical corpus and rebuild the retrieval index. In practice, the dominant system cost remains nearest-neighbor indexing and search (often accelerated by approximate nearest-neighbor libraries such as FAISS), which is unchanged by PEARL; PEARL simply supplies a better geometry for that search. The overall procedure first computes normalized class prototypes from labeled data, then trains $\phi_\theta$ using the combined alignment, contrast, and preservation objectives, and finally outputs enhanced embeddings $\tilde{\mathbf{x}}_i = \phi_\theta(\mathbf{x}_i)$ for downstream indexing and retrieval.

Finally, \method\ is not simply centering, L2 normalization, or whitening---those operations modify global statistics without directly optimizing neighborhood-label consistency. In contrast, \method\ uses limited label information to actively reshape semantic neighborhoods around class prototypes. Normalization remains important for retrieval stability, as most cosine-based retrieval pipelines L2-normalize embeddings before building an index. \method\ complements this: rather than replacing normalization, it provides a supervised geometry correction before (or alongside) standard normalization steps. Conceptually, whitening aims to improve \emph{global} isotropy, while \method\ aims to improve \emph{label-consistent local neighborhoods}. The results in \Cref{tab:purity,tab:hit,tab:mrr} show that these goals can differ: a method can improve Purity@K and Hit@1 while another method improves broader Hit@K at larger $K$.

\section{Experimental Setup}
\label{sec:setup}

We evaluate \method\ on embeddings from a digital governance communication corpus involving citizen messages and related governance categories. We use stratified cross-validation (five folds in our implementation), and to model label scarcity, for each fold we sample a roughly class-balanced labeled training subset with size ranging from 100 up to 5000 (the specific budgets are reported in the result tables), reserving a small validation split (15\% in our implementation) for early stopping and model selection. Each instance is a text message (or short document) produced in a citizen--government interaction channel. The downstream goal is to support retrieval and triage: given a new message, retrieve similar historical cases and infer a category label that is actionable for routing and handling. This setting naturally prioritizes \emph{early ranking quality} because human operators typically inspect only the first few retrieved cases.

We assume a fixed encoder $f(\cdot)$ maps text to embeddings $\mathbf{x} = f(x) \in \R^d$. The encoder is held fixed across all experiments to reflect the common deployment constraint that the embedding service cannot be fine-tuned for every downstream use case. For retrieval, we use cosine similarity
\begin{equation}
    \mathrm{sim}(\mathbf{x},\mathbf{x}') = \frac{\mathbf{x}^\top \mathbf{x}'}{\|\mathbf{x}\|_2\|\mathbf{x}'\|_2},
\end{equation}
and nearest-neighbor search is performed under cosine distance (equivalently cosine similarity), which is scale-invariant and therefore implicitly normalizes vector norms. Before training PEARL or fitting baselines, we standardize embeddings using z-score normalization (StandardScaler) fitted on the training split and applied to validation and test. To evaluate label efficiency, we vary the number of labeled training points available for building prototypes and training refinement. For each training size, we repeat the sampling procedure across multiple random seeds (or folds) and report mean $\pm$ standard deviation, matching the reporting in \Cref{tab:purity,tab:delta,tab:hit,tab:mrr}. This protocol reflects what matters operationally: performance variability can be as important as mean performance when labels are scarce.

We compare \method\ to representative embedding post-processing and supervised projection baselines: Raw uses the original embeddings, L2 applies L2 normalization, PCA-whiten+L2 applies whitening followed by L2 normalization \citep{su2021whitening}, and LDA+L2 applies a supervised linear projection followed by L2 normalization. These baselines cover three common strategies used in practice: \textbf{Raw} and \textbf{L2} represent minimal processing that many systems deploy by default; \textbf{PCA-whiten+L2} represents unsupervised geometry correction that can improve isotropy and retrieval speed \citep{su2021whitening}; and \textbf{LDA+L2} represents a strongly supervised approach that can be highly effective when enough labeled data exists, but can become unstable when labels are scarce. Importantly, these baselines differ in the kinds of behaviors they favor: whitening tends to improve broader recall at higher $K$ by flattening variance structure, while supervised projections may optimize discriminability in ways that reshape global structure. \method\ is designed to target \emph{local} neighborhoods and early ranking while remaining stable when labels are scarce.

We report metrics that directly reflect neighborhood structure and retrieval quality: neighborhood quality (Purity@K and intra--inter similarity separation $\Delta_{\mathrm{sep}}$), retrieval effectiveness (Hit@K and MRR@K, emphasizing early ranking behavior), and kNN-based classification as a lightweight label-efficient classifier. We use standard information retrieval metrics for ranked neighbor lists \citep{manning2008ir}, alongside geometry diagnostics that directly test label consistency in local neighborhoods. To connect with contemporary IR evaluation practice, these metrics are aligned with the broader use of benchmark suites for assessing robustness of embedding-based retrieval across domains \citep{thakur2021beir}.

For neighborhood label purity (Purity@K), for each query point $i$ with label $y_i$, let $\mathcal{N}_K(i)$ be its $K$ nearest neighbors in the training set (under cosine similarity). Purity@K is the fraction of neighbors that share the query label:
\begin{equation}
    \mathrm{Purity@}K = \frac{1}{|\mathcal{Q}|}\sum_{i \in \mathcal{Q}} \frac{1}{K}\sum_{j \in \mathcal{N}_K(i)} \mathbb{I}[y_j = y_i].
\end{equation}
This metric directly tests whether the embedding space supports label-consistent neighborhoods, which is a prerequisite for reliable kNN-style inference.

To measure class separation, we compute the gap between average intra-class similarity and average inter-class similarity:
\begin{equation}
    \Delta_{\mathrm{sep}} = \E[\mathrm{sim}(\mathbf{x},\mathbf{x}') \mid y=y'] - \E[\mathrm{sim}(\mathbf{x},\mathbf{x}') \mid y\neq y'].
\end{equation}
Larger $\Delta_{\mathrm{sep}}$ indicates a geometry where same-class points are more aligned and different-class points are more separated on average.

For label-conditioned retrieval, Hit@K counts whether at least one of the top-$K$ retrieved neighbors has the same label as the query:
\begin{equation}
    \mathrm{Hit@}K = \frac{1}{|\mathcal{Q}|}\sum_{i \in \mathcal{Q}} \mathbb{I}\left[\exists j \in \mathcal{N}_K(i)\ \text{s.t.}\ y_j = y_i\right].
\end{equation}
This metric captures recall-style behavior: it asks whether a correct neighbor is present somewhere in the top-$K$ list.

To evaluate early ranking quality, mean reciprocal rank (truncated at $K$) rewards methods that place a correct neighbor very early. Let $r_i$ be the rank position (1-indexed) of the first correct-label neighbor for query $i$ within the top-$K$ list, or $r_i=\infty$ if none exists. Then
\begin{equation}
    \mathrm{MRR@}K = \frac{1}{|\mathcal{Q}|}\sum_{i \in \mathcal{Q}} \mathbb{I}[r_i \le K]\cdot \frac{1}{r_i}.
\end{equation}
In governance-style workflows, where operators inspect the first few retrieved cases, MRR is often more indicative than Hit@K at large $K$.

Although our emphasis is on retrieval and geometry, it is useful to connect the metrics above to a simple label-efficient classifier that is common in practice. The kNN classifier \citep{cover1967nearest} predicts $\hat{y}_i$ by taking the majority label among $\mathcal{N}_K(i)$ (optionally weighted by similarity). This method inherits its behavior directly from the embedding geometry, which is why improvements in Purity@K, Hit@1, and MRR@K often translate into tangible gains for non-parametric decision support.

\section{Results}
\label{sec:results}

\method\ improves neighborhood quality and early retrieval (especially Hit@1 and MRR) in label-scarce and moderate regimes. As label availability increases, supervised projections (e.g., LDA+L2) can match or exceed \method, consistent with the goal of \method\ as a label-efficient geometry reshaping method rather than a replacement for fully supervised learning. Two high-level patterns recur across metrics: first, methods that explicitly use label information (PEARL and LDA+L2) tend to improve \emph{purity and separation} relative to raw embeddings, reflecting the fact that labels provide direct information about which neighborhoods should be consistent; second, the ranking metrics reveal a nuanced tradeoff where a method can be very strong at \emph{Hit@1} and MRR (early precision) while being less strong at \emph{Hit@}K for larger $K$ (recall-oriented retrieval). This is important for system design because different applications prioritize different parts of the ranked list. We emphasize that the goal of \method\ is not to win every metric at every training size, but to provide a robust improvement in the regime that dominates practice: low-label operation with fixed embeddings, where retrieval-based decisions are common and early ranking errors are costly.

In the low-label regime, with very limited labels (e.g., training size 100), \method\ provides strong improvements in Purity@K and Hit@1 compared to raw embeddings and common unsupervised post-processing (\Cref{tab:purity,tab:hit}). For broader recall-oriented retrieval (Hit@5/10/20), whitening can outperform, highlighting that different post-processing choices trade off early precision, recall, and neighborhood purity. In this regime, PEARL achieves the highest Purity@K for all reported $K$ values, indicating that it most effectively reshapes local neighborhoods to align with labels. At the same time, \Cref{tab:hit} shows that whitening can produce very high Hit@K at larger $K$. One interpretation is that whitening makes the space more isotropic and spreads points in a way that improves recall, but does not necessarily produce the most label-consistent \emph{nearest} neighbors. PEARL, in contrast, prioritizes neighborhood correctness at the top of the list, which is also reflected in improved Hit@1 and early ranking behavior.

As labeled size increases (e.g., 600--2500), \method\ remains competitive and continues to improve early ranking behavior (MRR@K), while LDA-based projections begin to catch up on some metrics (\Cref{tab:purity,tab:delta,tab:mrr}). At training sizes around 600 and 1200, PEARL continues to deliver strong Hit@1 and often improves MRR@K compared to raw, L2, and whitening baselines (\Cref{tab:mrr}), indicating that PEARL not only increases the chance of retrieving a correct neighbor, but also tends to rank correct neighbors earlier. Meanwhile, the purity and separation metrics show that supervised projections can become highly effective as labels increase, especially when the projection has enough labeled support to estimate class statistics reliably. This regime is particularly informative for practice: many deployments can afford hundreds to a few thousand labels, but not tens of thousands. In that range, PEARL offers a simple way to improve retrieval quality without committing to a full supervised projection pipeline.

At higher labeled sizes (e.g., 5000), LDA+L2 becomes the strongest performer on several geometry and retrieval metrics, while \method\ remains consistently better than raw embeddings, confirming that \method\ is most valuable in the label-scarce setting that motivates the method. As labels become plentiful, supervised projections can exploit richer class statistics to maximize separation and retrieval performance, and the advantage of PEARL's label-efficient inductive bias naturally shrinks. Importantly, PEARL does not degrade performance relative to the raw baseline in this regime; instead, it continues to provide a competitive alternative when full supervision is not desirable, or when a system requires a lightweight preprocessing step that is easy to deploy and maintain.

\begin{table}[t]
\centering
\caption{Neighbor label purity (mean $\pm$ std) under cosine retrieval. Higher is better.}
\label{tab:purity}
\small
\resizebox{\textwidth}{!}{%
\begin{tabular}{rrlllll}
\toprule
Train size & k & Raw & PEARL & L2 & PCA-whiten+L2 & LDA+L2 \\
\midrule
100 & 1 & $0.3386 \pm 0.0211$ & \boldmath\textbf{$0.4256 \pm 0.0230$} & $0.3386 \pm 0.0211$ & $0.3514 \pm 0.0171$ & $0.3510 \pm 0.0188$ \\
 & 5 & $0.2439 \pm 0.0068$ & \boldmath\textbf{$0.3782 \pm 0.0234$} & $0.2439 \pm 0.0068$ & $0.2373 \pm 0.0087$ & $0.3006 \pm 0.0194$ \\
 & 10 & $0.1942 \pm 0.0059$ & \boldmath\textbf{$0.2860 \pm 0.0110$} & $0.1942 \pm 0.0059$ & $0.1846 \pm 0.0056$ & $0.2446 \pm 0.0115$ \\
 & 20 & $0.1502 \pm 0.0027$ & \boldmath\textbf{$0.1946 \pm 0.0069$} & $0.1502 \pm 0.0027$ & $0.1427 \pm 0.0034$ & $0.1804 \pm 0.0061$ \\
\addlinespace
300 & 1 & $0.3824 \pm 0.0053$ & $0.5370 \pm 0.0544$ & $0.3824 \pm 0.0053$ & $0.3388 \pm 0.0219$ & \boldmath\textbf{$0.5932 \pm 0.0169$} \\
 & 5 & $0.3086 \pm 0.0114$ & $0.4802 \pm 0.0527$ & $0.3086 \pm 0.0114$ & $0.2393 \pm 0.0088$ & \boldmath\textbf{$0.5897 \pm 0.0179$} \\
 & 10 & $0.2690 \pm 0.0079$ & $0.4304 \pm 0.0504$ & $0.2690 \pm 0.0079$ & $0.1966 \pm 0.0051$ & \boldmath\textbf{$0.5810 \pm 0.0134$} \\
 & 20 & $0.2242 \pm 0.0049$ & $0.3518 \pm 0.0478$ & $0.2242 \pm 0.0049$ & $0.1610 \pm 0.0037$ & \boldmath\textbf{$0.5343 \pm 0.0132$} \\
\addlinespace
600 & 1 & $0.4386 \pm 0.0217$ & \boldmath\textbf{$0.6054 \pm 0.0290$} & $0.4386 \pm 0.0217$ & $0.4422 \pm 0.0123$ & $0.5824 \pm 0.0140$ \\
 & 5 & $0.3688 \pm 0.0139$ & $0.5402 \pm 0.0393$ & $0.3688 \pm 0.0139$ & $0.3381 \pm 0.0059$ & \boldmath\textbf{$0.5812 \pm 0.0092$} \\
 & 10 & $0.3268 \pm 0.0076$ & $0.4854 \pm 0.0440$ & $0.3268 \pm 0.0076$ & $0.2827 \pm 0.0031$ & \boldmath\textbf{$0.5809 \pm 0.0081$} \\
 & 20 & $0.2799 \pm 0.0054$ & $0.4107 \pm 0.0452$ & $0.2799 \pm 0.0054$ & $0.2264 \pm 0.0029$ & \boldmath\textbf{$0.5777 \pm 0.0080$} \\
\addlinespace
1200 & 1 & $0.4742 \pm 0.0137$ & \boldmath\textbf{$0.6348 \pm 0.0333$} & $0.4742 \pm 0.0137$ & $0.4900 \pm 0.0150$ & $0.5172 \pm 0.0165$ \\
 & 5 & $0.4090 \pm 0.0032$ & \boldmath\textbf{$0.5784 \pm 0.0431$} & $0.4090 \pm 0.0031$ & $0.4007 \pm 0.0054$ & $0.5156 \pm 0.0149$ \\
 & 10 & $0.3684 \pm 0.0028$ & \boldmath\textbf{$0.5387 \pm 0.0535$} & $0.3684 \pm 0.0028$ & $0.3485 \pm 0.0049$ & $0.5159 \pm 0.0164$ \\
 & 20 & $0.3238 \pm 0.0029$ & $0.4802 \pm 0.0612$ & $0.3238 \pm 0.0029$ & $0.2907 \pm 0.0024$ & \boldmath\textbf{$0.5151 \pm 0.0169$} \\
\addlinespace
2500 & 1 & $0.5150 \pm 0.0233$ & $0.6826 \pm 0.0456$ & $0.5150 \pm 0.0233$ & $0.5198 \pm 0.0190$ & \boldmath\textbf{$0.7378 \pm 0.0166$} \\
 & 5 & $0.4540 \pm 0.0127$ & $0.6359 \pm 0.0671$ & $0.4540 \pm 0.0127$ & $0.4527 \pm 0.0098$ & \boldmath\textbf{$0.7336 \pm 0.0140$} \\
 & 10 & $0.4154 \pm 0.0060$ & $0.6022 \pm 0.0779$ & $0.4154 \pm 0.0060$ & $0.4108 \pm 0.0052$ & \boldmath\textbf{$0.7315 \pm 0.0143$} \\
 & 20 & $0.3736 \pm 0.0043$ & $0.5612 \pm 0.0886$ & $0.3736 \pm 0.0043$ & $0.3582 \pm 0.0039$ & \boldmath\textbf{$0.7289 \pm 0.0152$} \\
\addlinespace
5000 & 1 & $0.5400 \pm 0.0160$ & $0.6648 \pm 0.0397$ & $0.5400 \pm 0.0160$ & $0.5366 \pm 0.0216$ & \boldmath\textbf{$0.7622 \pm 0.0137$} \\
 & 5 & $0.4812 \pm 0.0090$ & $0.6216 \pm 0.0519$ & $0.4812 \pm 0.0090$ & $0.4779 \pm 0.0103$ & \boldmath\textbf{$0.7620 \pm 0.0140$} \\
 & 10 & $0.4429 \pm 0.0053$ & $0.5870 \pm 0.0558$ & $0.4429 \pm 0.0053$ & $0.4414 \pm 0.0083$ & \boldmath\textbf{$0.7618 \pm 0.0136$} \\
 & 20 & $0.4006 \pm 0.0038$ & $0.5463 \pm 0.0626$ & $0.4006 \pm 0.0038$ & $0.3945 \pm 0.0066$ & \boldmath\textbf{$0.7602 \pm 0.0134$} \\
\bottomrule
\end{tabular}%
}
\end{table}

The purity results highlight the core strength of PEARL: in the most label-scarce setting (training size 100), PEARL yields markedly higher label-consistency among nearest neighbors across all $K$. As labels increase, LDA+L2 becomes dominant, especially at larger $K$, reflecting the fact that a well-estimated supervised projection can strongly enforce class separation. Nevertheless, PEARL remains competitive and often improves over raw embeddings, suggesting that prototype-based alignment is a robust and practical approach for local neighborhood correction.

\begin{table}[t]
\centering
\caption{Similarity separation $\Delta_{\mathrm{sep}}$ (intra-class minus inter-class cosine similarity; mean $\pm$ std). Higher is better.}
\label{tab:delta}
\small
\resizebox{\textwidth}{!}{%
\begin{tabular}{rlllll}
\toprule
Train size & Raw & PEARL & L2 & PCA-whiten+L2 & LDA+L2 \\
\midrule
100 & $0.1226 \pm 0.0028$ & \boldmath\textbf{$0.3496 \pm 0.0278$} & $0.1226 \pm 0.0028$ & $0.1125 \pm 0.0024$ & $0.2778 \pm 0.0137$ \\
\addlinespace
300 & $0.1251 \pm 0.0048$ & $0.3690 \pm 0.0817$ & $0.1251 \pm 0.0048$ & $0.0468 \pm 0.0025$ & \boldmath\textbf{$0.4346 \pm 0.0165$} \\
\addlinespace
600 & $0.1232 \pm 0.0028$ & $0.3020 \pm 0.0475$ & $0.1232 \pm 0.0028$ & $0.0396 \pm 0.0007$ & \boldmath\textbf{$0.4714 \pm 0.0068$} \\
\addlinespace
1200 & $0.1211 \pm 0.0038$ & $0.3102 \pm 0.0723$ & $0.1211 \pm 0.0038$ & $0.0360 \pm 0.0011$ & \boldmath\textbf{$0.5342 \pm 0.0086$} \\
\addlinespace
2500 & $0.1176 \pm 0.0057$ & $0.3335 \pm 0.0991$ & $0.1176 \pm 0.0057$ & $0.0326 \pm 0.0017$ & \boldmath\textbf{$0.6916 \pm 0.0067$} \\
\addlinespace
5000 & $0.1155 \pm 0.0022$ & $0.2732 \pm 0.0643$ & $0.1155 \pm 0.0022$ & $0.0297 \pm 0.0008$ & \boldmath\textbf{$0.6992 \pm 0.0046$} \\
\bottomrule
\end{tabular}%
}
\end{table}

The similarity separation metric provides a complementary global view of geometry. PEARL substantially increases separation at training size 100, consistent with the idea that prototype alignment increases intra-class similarity while pushing away from other prototypes. At larger training sizes, LDA+L2 yields the highest separation, which aligns with its objective of maximizing class discrimination when labels are sufficiently abundant.

\begin{table}[t]
\centering
\caption{Retrieval Hit@K (mean $\pm$ std). Higher is better.}
\label{tab:hit}
\small
\resizebox{\textwidth}{!}{%
\begin{tabular}{rrlllll}
\toprule
Train size & k & Raw & PEARL & L2 & PCA-whiten+L2 & LDA+L2 \\
\midrule
100 & 1 & $0.3386 \pm 0.0211$ & \boldmath\textbf{$0.4256 \pm 0.0230$} & $0.3386 \pm 0.0211$ & $0.3514 \pm 0.0171$ & $0.3510 \pm 0.0188$ \\
 & 5 & $0.7092 \pm 0.0146$ & $0.5834 \pm 0.0324$ & $0.7092 \pm 0.0146$ & \boldmath\textbf{$0.7516 \pm 0.0206$} & $0.6762 \pm 0.0182$ \\
 & 10 & $0.8638 \pm 0.0189$ & $0.7182 \pm 0.0405$ & $0.8638 \pm 0.0189$ & \boldmath\textbf{$0.8938 \pm 0.0119$} & $0.8342 \pm 0.0082$ \\
 & 20 & $0.9694 \pm 0.0067$ & $0.8608 \pm 0.0345$ & $0.9694 \pm 0.0067$ & \boldmath\textbf{$0.9810 \pm 0.0073$} & $0.9420 \pm 0.0062$ \\
\addlinespace
300 & 1 & $0.3824 \pm 0.0053$ & $0.5370 \pm 0.0544$ & $0.3824 \pm 0.0053$ & $0.3388 \pm 0.0219$ & \boldmath\textbf{$0.5932 \pm 0.0169$} \\
 & 5 & \boldmath\textbf{$0.7518 \pm 0.0137$} & $0.7214 \pm 0.0432$ & \boldmath\textbf{$0.7518 \pm 0.0137$} & $0.7300 \pm 0.0189$ & $0.7500 \pm 0.0144$ \\
 & 10 & \boldmath\textbf{$0.8856 \pm 0.0104$} & $0.8058 \pm 0.0527$ & \boldmath\textbf{$0.8856 \pm 0.0104$} & $0.8784 \pm 0.0119$ & $0.8054 \pm 0.0085$ \\
 & 20 & $0.9622 \pm 0.0023$ & $0.8914 \pm 0.0463$ & $0.9622 \pm 0.0023$ & \boldmath\textbf{$0.9674 \pm 0.0072$} & $0.8810 \pm 0.0045$ \\
\addlinespace
600 & 1 & $0.4386 \pm 0.0217$ & \boldmath\textbf{$0.6054 \pm 0.0290$} & $0.4386 \pm 0.0217$ & $0.4422 \pm 0.0123$ & $0.5824 \pm 0.0140$ \\
 & 5 & $0.7950 \pm 0.0230$ & \boldmath\textbf{$0.8008 \pm 0.0163$} & $0.7950 \pm 0.0230$ & $0.7986 \pm 0.0114$ & $0.6748 \pm 0.0101$ \\
 & 10 & $0.9070 \pm 0.0041$ & $0.8724 \pm 0.0168$ & $0.9070 \pm 0.0041$ & \boldmath\textbf{$0.9186 \pm 0.0069$} & $0.7110 \pm 0.0120$ \\
 & 20 & $0.9660 \pm 0.0040$ & $0.9354 \pm 0.0168$ & $0.9660 \pm 0.0040$ & \boldmath\textbf{$0.9792 \pm 0.0040$} & $0.7540 \pm 0.0124$ \\
\addlinespace
1200 & 1 & $0.4742 \pm 0.0137$ & \boldmath\textbf{$0.6348 \pm 0.0333$} & $0.4742 \pm 0.0137$ & $0.4900 \pm 0.0150$ & $0.5172 \pm 0.0165$ \\
 & 5 & $0.8146 \pm 0.0080$ & $0.8254 \pm 0.0098$ & $0.8146 \pm 0.0080$ & \boldmath\textbf{$0.8342 \pm 0.0126$} & $0.5872 \pm 0.0144$ \\
 & 10 & $0.9148 \pm 0.0047$ & $0.8904 \pm 0.0123$ & $0.9148 \pm 0.0047$ & \boldmath\textbf{$0.9308 \pm 0.0094$} & $0.6192 \pm 0.0167$ \\
 & 20 & $0.9686 \pm 0.0032$ & $0.9410 \pm 0.0104$ & $0.9686 \pm 0.0032$ & \boldmath\textbf{$0.9844 \pm 0.0036$} & $0.6546 \pm 0.0150$ \\
\addlinespace
2500 & 1 & $0.5150 \pm 0.0233$ & $0.6826 \pm 0.0456$ & $0.5150 \pm 0.0233$ & $0.5198 \pm 0.0190$ & \boldmath\textbf{$0.7378 \pm 0.0166$} \\
 & 5 & $0.8372 \pm 0.0188$ & \boldmath\textbf{$0.8542 \pm 0.0149$} & $0.8372 \pm 0.0188$ & $0.8528 \pm 0.0128$ & $0.8278 \pm 0.0114$ \\
 & 10 & $0.9242 \pm 0.0106$ & $0.9104 \pm 0.0113$ & $0.9242 \pm 0.0106$ & \boldmath\textbf{$0.9420 \pm 0.0042$} & $0.8544 \pm 0.0138$ \\
 & 20 & $0.9712 \pm 0.0054$ & $0.9474 \pm 0.0086$ & $0.9712 \pm 0.0054$ & \boldmath\textbf{$0.9846 \pm 0.0017$} & $0.8808 \pm 0.0081$ \\
\addlinespace
5000 & 1 & $0.5400 \pm 0.0160$ & $0.6648 \pm 0.0397$ & $0.5400 \pm 0.0160$ & $0.5366 \pm 0.0216$ & \boldmath\textbf{$0.7622 \pm 0.0137$} \\
 & 5 & $0.8498 \pm 0.0128$ & $0.8564 \pm 0.0141$ & $0.8498 \pm 0.0128$ & $0.8594 \pm 0.0133$ & \boldmath\textbf{$0.8668 \pm 0.0116$} \\
 & 10 & $0.9282 \pm 0.0088$ & $0.9218 \pm 0.0080$ & $0.9282 \pm 0.0088$ & \boldmath\textbf{$0.9430 \pm 0.0037$} & $0.8920 \pm 0.0065$ \\
 & 20 & $0.9728 \pm 0.0053$ & $0.9596 \pm 0.0070$ & $0.9728 \pm 0.0053$ & \boldmath\textbf{$0.9834 \pm 0.0062$} & $0.9150 \pm 0.0043$ \\
\bottomrule
\end{tabular}%
}
\end{table}

Hit@K illustrates why it is valuable to report both purity and ranking metrics. For $K=1$, PEARL often achieves the best or near-best performance, especially in the low-label and moderate-label regimes, indicating stronger \emph{top-1} retrieval. For larger $K$, whitening frequently provides the strongest results, suggesting that making the space more globally isotropic can improve recall-style retrieval even if the nearest neighbors are less label-pure. This difference matters in applications: if the system surfaces only a few cases to a human operator, PEARL's Hit@1 gains are particularly relevant.

\begin{table}[t]
\centering
\caption{Retrieval MRR@K (mean $\pm$ std). Higher is better.}
\label{tab:mrr}
\small
\resizebox{\textwidth}{!}{%
\begin{tabular}{rrlllll}
\toprule
Train size & k & Raw & PEARL & L2 & PCA-whiten+L2 & LDA+L2 \\
\midrule
100 & 1 & $0.3386 \pm 0.0211$ & \boldmath\textbf{$0.4256 \pm 0.0230$} & $0.3386 \pm 0.0211$ & $0.3514 \pm 0.0171$ & $0.3510 \pm 0.0188$ \\
 & 5 & $0.4727 \pm 0.0148$ & $0.4792 \pm 0.0238$ & $0.4727 \pm 0.0148$ & \boldmath\textbf{$0.4975 \pm 0.0132$} & $0.4684 \pm 0.0198$ \\
 & 10 & $0.4936 \pm 0.0154$ & $0.4972 \pm 0.0251$ & $0.4936 \pm 0.0154$ & \boldmath\textbf{$0.5169 \pm 0.0116$} & $0.4896 \pm 0.0185$ \\
 & 20 & $0.5013 \pm 0.0148$ & $0.5072 \pm 0.0245$ & $0.5013 \pm 0.0148$ & \boldmath\textbf{$0.5233 \pm 0.0115$} & $0.4974 \pm 0.0180$ \\
\addlinespace
300 & 1 & $0.3824 \pm 0.0053$ & $0.5370 \pm 0.0544$ & $0.3824 \pm 0.0053$ & $0.3388 \pm 0.0219$ & \boldmath\textbf{$0.5932 \pm 0.0169$} \\
 & 5 & $0.5193 \pm 0.0083$ & $0.6049 \pm 0.0463$ & $0.5193 \pm 0.0083$ & $0.4824 \pm 0.0195$ & \boldmath\textbf{$0.6530 \pm 0.0149$} \\
 & 10 & $0.5374 \pm 0.0076$ & $0.6164 \pm 0.0460$ & $0.5374 \pm 0.0076$ & $0.5026 \pm 0.0187$ & \boldmath\textbf{$0.6604 \pm 0.0138$} \\
 & 20 & $0.5429 \pm 0.0069$ & $0.6224 \pm 0.0456$ & $0.5429 \pm 0.0069$ & $0.5090 \pm 0.0182$ & \boldmath\textbf{$0.6655 \pm 0.0138$} \\
\addlinespace
600 & 1 & $0.4386 \pm 0.0217$ & \boldmath\textbf{$0.6054 \pm 0.0290$} & $0.4386 \pm 0.0217$ & $0.4422 \pm 0.0123$ & $0.5824 \pm 0.0140$ \\
 & 5 & $0.5725 \pm 0.0207$ & \boldmath\textbf{$0.6790 \pm 0.0201$} & $0.5725 \pm 0.0207$ & $0.5757 \pm 0.0102$ & $0.6173 \pm 0.0122$ \\
 & 10 & $0.5877 \pm 0.0178$ & \boldmath\textbf{$0.6887 \pm 0.0201$} & $0.5877 \pm 0.0178$ & $0.5919 \pm 0.0096$ & $0.6220 \pm 0.0122$ \\
 & 20 & $0.5920 \pm 0.0178$ & \boldmath\textbf{$0.6932 \pm 0.0198$} & $0.5920 \pm 0.0178$ & $0.5964 \pm 0.0097$ & $0.6250 \pm 0.0123$ \\
\addlinespace
1200 & 1 & $0.4742 \pm 0.0137$ & \boldmath\textbf{$0.6348 \pm 0.0333$} & $0.4742 \pm 0.0137$ & $0.4900 \pm 0.0150$ & $0.5172 \pm 0.0165$ \\
 & 5 & $0.6025 \pm 0.0102$ & \boldmath\textbf{$0.7064 \pm 0.0186$} & $0.6025 \pm 0.0102$ & $0.6190 \pm 0.0099$ & $0.5441 \pm 0.0150$ \\
 & 10 & $0.6163 \pm 0.0101$ & \boldmath\textbf{$0.7151 \pm 0.0180$} & $0.6163 \pm 0.0101$ & $0.6323 \pm 0.0094$ & $0.5485 \pm 0.0153$ \\
 & 20 & $0.6202 \pm 0.0101$ & \boldmath\textbf{$0.7186 \pm 0.0182$} & $0.6202 \pm 0.0101$ & $0.6362 \pm 0.0093$ & $0.5509 \pm 0.0152$ \\
\addlinespace
2500 & 1 & $0.5150 \pm 0.0233$ & $0.6826 \pm 0.0456$ & $0.5150 \pm 0.0233$ & $0.5198 \pm 0.0190$ & \boldmath\textbf{$0.7378 \pm 0.0166$} \\
 & 5 & $0.6388 \pm 0.0211$ & $0.7471 \pm 0.0332$ & $0.6388 \pm 0.0211$ & $0.6461 \pm 0.0163$ & \boldmath\textbf{$0.7735 \pm 0.0134$} \\
 & 10 & $0.6508 \pm 0.0194$ & $0.7547 \pm 0.0313$ & $0.6508 \pm 0.0194$ & $0.6582 \pm 0.0151$ & \boldmath\textbf{$0.7771 \pm 0.0136$} \\
 & 20 & $0.6542 \pm 0.0195$ & $0.7573 \pm 0.0311$ & $0.6542 \pm 0.0195$ & $0.6613 \pm 0.0149$ & \boldmath\textbf{$0.7789 \pm 0.0133$} \\
\addlinespace
5000 & 1 & $0.5400 \pm 0.0160$ & $0.6648 \pm 0.0397$ & $0.5400 \pm 0.0160$ & $0.5366 \pm 0.0216$ & \boldmath\textbf{$0.7622 \pm 0.0137$} \\
 & 5 & $0.6603 \pm 0.0149$ & $0.7400 \pm 0.0284$ & $0.6603 \pm 0.0149$ & $0.6598 \pm 0.0181$ & \boldmath\textbf{$0.8042 \pm 0.0121$} \\
 & 10 & $0.6710 \pm 0.0141$ & $0.7486 \pm 0.0265$ & $0.6710 \pm 0.0141$ & $0.6714 \pm 0.0166$ & \boldmath\textbf{$0.8077 \pm 0.0112$} \\
 & 20 & $0.6742 \pm 0.0138$ & $0.7513 \pm 0.0263$ & $0.6742 \pm 0.0138$ & $0.6744 \pm 0.0165$ & \boldmath\textbf{$0.8093 \pm 0.0109$} \\
\bottomrule
\end{tabular}%
}
\end{table}

MRR@K further strengthens the early-ranking story. Improvements in MRR indicate that correct-label neighbors not only exist within the top-$K$ list but tend to appear near the top. In several moderate-label settings, PEARL yields strong MRR@K improvements, supporting its role as an early-precision enhancement method. In higher-label settings, LDA+L2 often dominates MRR as well, consistent with its ability to strongly optimize class separation when sufficient supervision is available.

To directly connect geometry improvements to a lightweight classification task, we report kNN F1 scores for a representative class (Class 1) under two common kNN variants: uniform voting (\Cref{tab:knn_f1_uniform}) and distance-weighted voting (\Cref{tab:knn_f1_distance}). These results mirror the retrieval findings: PEARL often improves over Raw/L2 and is especially helpful in low-label regimes, while strong supervised projections (LDA+L2) can become competitive or dominant as label availability increases.

\begin{table}[t]
\centering
\caption{Class 1 F1 for kNN with uniform voting (mean $\pm$ std). Higher is better.}
\label{tab:knn_f1_uniform}
\small
\resizebox{\textwidth}{!}{%
\begin{tabular}{rrlllll}
\toprule
Train size & k & Raw & PEARL & L2 & PCA-whiten+L2 & LDA+L2 \\
\midrule
100 & 1 & $0.3122 \pm 0.0856$ & \boldmath\textbf{$0.4125 \pm 0.0428$} & $0.3122 \pm 0.0856$ & $0.2886 \pm 0.0753$ & $0.3372 \pm 0.0613$ \\
 & 5 & $0.3321 \pm 0.0295$ & \boldmath\textbf{$0.4071 \pm 0.0292$} & $0.3321 \pm 0.0295$ & $0.2979 \pm 0.0443$ & $0.3441 \pm 0.0620$ \\
 & 10 & $0.3423 \pm 0.0278$ & \boldmath\textbf{$0.3968 \pm 0.0390$} & $0.3423 \pm 0.0278$ & $0.3433 \pm 0.0303$ & $0.3746 \pm 0.0523$ \\
 & 20 & $0.3218 \pm 0.0131$ & $0.3760 \pm 0.0454$ & $0.3218 \pm 0.0131$ & $0.3433 \pm 0.0321$ & \boldmath\textbf{$0.3899 \pm 0.0393$} \\
\addlinespace
300 & 1 & $0.3114 \pm 0.0588$ & $0.4566 \pm 0.0345$ & $0.3114 \pm 0.0588$ & $0.3070 \pm 0.0471$ & \boldmath\textbf{$0.4604 \pm 0.0635$} \\
 & 5 & $0.3324 \pm 0.0663$ & $0.4570 \pm 0.0275$ & $0.3324 \pm 0.0663$ & $0.2957 \pm 0.0295$ & \boldmath\textbf{$0.4946 \pm 0.0523$} \\
 & 10 & $0.3818 \pm 0.0586$ & $0.4542 \pm 0.0400$ & $0.3818 \pm 0.0586$ & $0.3330 \pm 0.0529$ & \boldmath\textbf{$0.4955 \pm 0.0838$} \\
 & 20 & $0.3816 \pm 0.0338$ & $0.3969 \pm 0.0553$ & $0.3816 \pm 0.0338$ & $0.3217 \pm 0.0461$ & \boldmath\textbf{$0.4835 \pm 0.0595$} \\
\addlinespace
600 & 1 & $0.3503 \pm 0.0482$ & \boldmath\textbf{$0.4966 \pm 0.0490$} & $0.3503 \pm 0.0482$ & $0.3599 \pm 0.0536$ & $0.4195 \pm 0.0754$ \\
 & 5 & $0.3910 \pm 0.0284$ & \boldmath\textbf{$0.4900 \pm 0.0468$} & $0.3910 \pm 0.0284$ & $0.3775 \pm 0.0385$ & $0.4355 \pm 0.0726$ \\
 & 10 & $0.4267 \pm 0.0324$ & \boldmath\textbf{$0.4883 \pm 0.0480$} & $0.4267 \pm 0.0324$ & $0.4262 \pm 0.0259$ & $0.4404 \pm 0.0784$ \\
 & 20 & $0.4556 \pm 0.0493$ & \boldmath\textbf{$0.4754 \pm 0.0507$} & $0.4556 \pm 0.0493$ & $0.4377 \pm 0.0668$ & $0.4371 \pm 0.0754$ \\
\addlinespace
1200 & 1 & $0.3735 \pm 0.0310$ & \boldmath\textbf{$0.5645 \pm 0.0401$} & $0.3735 \pm 0.0310$ & $0.4196 \pm 0.0309$ & $0.3439 \pm 0.0561$ \\
 & 5 & $0.4315 \pm 0.0214$ & \boldmath\textbf{$0.5256 \pm 0.0338$} & $0.4315 \pm 0.0214$ & $0.4063 \pm 0.0413$ & $0.3440 \pm 0.0663$ \\
 & 10 & $0.4652 \pm 0.0139$ & \boldmath\textbf{$0.5393 \pm 0.0250$} & $0.4652 \pm 0.0139$ & $0.5074 \pm 0.0442$ & $0.3499 \pm 0.0783$ \\
 & 20 & $0.5073 \pm 0.0144$ & $0.5140 \pm 0.0482$ & $0.5073 \pm 0.0144$ & \boldmath\textbf{$0.5438 \pm 0.0245$} & $0.3548 \pm 0.0727$ \\
\addlinespace
2500 & 1 & $0.4030 \pm 0.0383$ & $0.5752 \pm 0.0787$ & $0.4030 \pm 0.0383$ & $0.4625 \pm 0.0279$ & \boldmath\textbf{$0.6144 \pm 0.0264$} \\
 & 5 & $0.4790 \pm 0.0183$ & $0.5836 \pm 0.0549$ & $0.4790 \pm 0.0183$ & $0.4858 \pm 0.0345$ & \boldmath\textbf{$0.6291 \pm 0.0374$} \\
 & 10 & $0.5240 \pm 0.0448$ & $0.5929 \pm 0.0626$ & $0.5240 \pm 0.0448$ & $0.5411 \pm 0.0432$ & \boldmath\textbf{$0.6267 \pm 0.0277$} \\
 & 20 & $0.5369 \pm 0.0357$ & $0.5892 \pm 0.0625$ & $0.5369 \pm 0.0357$ & $0.5675 \pm 0.0354$ & \boldmath\textbf{$0.6278 \pm 0.0348$} \\
\addlinespace
5000 & 1 & $0.4221 \pm 0.0429$ & $0.5760 \pm 0.0350$ & $0.4221 \pm 0.0429$ & $0.4476 \pm 0.0289$ & \boldmath\textbf{$0.6647 \pm 0.0429$} \\
 & 5 & $0.4933 \pm 0.0465$ & $0.5955 \pm 0.0230$ & $0.4933 \pm 0.0465$ & $0.4984 \pm 0.0282$ & \boldmath\textbf{$0.6828 \pm 0.0246$} \\
 & 10 & $0.5354 \pm 0.0565$ & $0.6074 \pm 0.0293$ & $0.5354 \pm 0.0565$ & $0.5323 \pm 0.0474$ & \boldmath\textbf{$0.6751 \pm 0.0193$} \\
 & 20 & $0.5615 \pm 0.0526$ & $0.6041 \pm 0.0270$ & $0.5615 \pm 0.0526$ & $0.5917 \pm 0.0316$ & \boldmath\textbf{$0.6701 \pm 0.0317$} \\
\bottomrule
\end{tabular}%
}
\end{table}

\begin{table}[t]
\centering
\caption{Class 1 F1 for kNN with distance-weighted voting (mean $\pm$ std). Higher is better.}
\label{tab:knn_f1_distance}
\small
\resizebox{\textwidth}{!}{%
\begin{tabular}{rrlllll}
\toprule
Train size & k & Raw & PEARL & L2 & PCA-whiten+L2 & LDA+L2 \\
\midrule
100 & 5 & $0.3199 \pm 0.0619$ & \boldmath\textbf{$0.4108 \pm 0.0324$} & $0.3199 \pm 0.0619$ & $0.3534 \pm 0.0672$ & $0.3485 \pm 0.0829$ \\
 & 10 & $0.3608 \pm 0.0615$ & \boldmath\textbf{$0.4104 \pm 0.0315$} & $0.3608 \pm 0.0615$ & $0.3877 \pm 0.0282$ & $0.3733 \pm 0.0798$ \\
 & 20 & $0.3645 \pm 0.0286$ & \boldmath\textbf{$0.4139 \pm 0.0350$} & $0.3645 \pm 0.0286$ & $0.3968 \pm 0.0342$ & $0.3852 \pm 0.0719$ \\
\addlinespace
300 & 5 & $0.3400 \pm 0.1010$ & $0.4608 \pm 0.0387$ & $0.3400 \pm 0.1010$ & $0.3165 \pm 0.0570$ & \boldmath\textbf{$0.4977 \pm 0.0556$} \\
 & 10 & $0.3767 \pm 0.0716$ & $0.4602 \pm 0.0350$ & $0.3767 \pm 0.0716$ & $0.3639 \pm 0.0666$ & \boldmath\textbf{$0.4894 \pm 0.0696$} \\
 & 20 & $0.3695 \pm 0.0377$ & $0.4569 \pm 0.0181$ & $0.3695 \pm 0.0377$ & $0.3513 \pm 0.0603$ & \boldmath\textbf{$0.4872 \pm 0.0571$} \\
\addlinespace
600 & 5 & $0.4078 \pm 0.0317$ & \boldmath\textbf{$0.5065 \pm 0.0600$} & $0.4078 \pm 0.0317$ & $0.4162 \pm 0.0511$ & $0.4291 \pm 0.0704$ \\
 & 10 & $0.4290 \pm 0.0230$ & \boldmath\textbf{$0.5051 \pm 0.0569$} & $0.4290 \pm 0.0230$ & $0.4472 \pm 0.0269$ & $0.4320 \pm 0.0790$ \\
 & 20 & $0.4789 \pm 0.0565$ & \boldmath\textbf{$0.4902 \pm 0.0641$} & $0.4789 \pm 0.0565$ & $0.4605 \pm 0.0579$ & $0.4351 \pm 0.0757$ \\
\addlinespace
1200 & 5 & $0.4621 \pm 0.0128$ & \boldmath\textbf{$0.5400 \pm 0.0281$} & $0.4621 \pm 0.0128$ & $0.4574 \pm 0.0428$ & $0.3446 \pm 0.0656$ \\
 & 10 & $0.4775 \pm 0.0113$ & \boldmath\textbf{$0.5700 \pm 0.0292$} & $0.4775 \pm 0.0113$ & $0.5145 \pm 0.0155$ & $0.3501 \pm 0.0780$ \\
 & 20 & $0.5104 \pm 0.0179$ & $0.5374 \pm 0.0273$ & $0.5104 \pm 0.0179$ & \boldmath\textbf{$0.5617 \pm 0.0393$} & $0.3545 \pm 0.0691$ \\
\addlinespace
2500 & 5 & $0.4720 \pm 0.0272$ & $0.5911 \pm 0.0501$ & $0.4720 \pm 0.0272$ & $0.5184 \pm 0.0373$ & \boldmath\textbf{$0.6297 \pm 0.0304$} \\
 & 10 & $0.5255 \pm 0.0585$ & $0.5990 \pm 0.0717$ & $0.5255 \pm 0.0585$ & $0.5385 \pm 0.0390$ & \boldmath\textbf{$0.6262 \pm 0.0296$} \\
 & 20 & $0.5426 \pm 0.0434$ & $0.5926 \pm 0.0652$ & $0.5426 \pm 0.0434$ & $0.5772 \pm 0.0360$ & \boldmath\textbf{$0.6301 \pm 0.0380$} \\
\addlinespace
5000 & 5 & $0.5195 \pm 0.0419$ & $0.6087 \pm 0.0355$ & $0.5195 \pm 0.0419$ & $0.5236 \pm 0.0382$ & \boldmath\textbf{$0.6845 \pm 0.0332$} \\
 & 10 & $0.5491 \pm 0.0549$ & $0.6114 \pm 0.0260$ & $0.5491 \pm 0.0549$ & $0.5624 \pm 0.0515$ & \boldmath\textbf{$0.6770 \pm 0.0191$} \\
 & 20 & $0.5622 \pm 0.0520$ & $0.6141 \pm 0.0241$ & $0.5622 \pm 0.0520$ & $0.6057 \pm 0.0483$ & \boldmath\textbf{$0.6708 \pm 0.0309$} \\
\bottomrule
\end{tabular}%
}
\end{table}

\section{Analysis and Interpretation}
\label{sec:analysis}

Across label-scarce settings, \method\ improves local neighborhood geometry: it increases neighbor-label consistency (Purity@K) and improves early ranking behavior (Hit@1 and MRR@K). This section interprets these effects through the lens of prototype-guided geometry shaping. First, prototypes provide low-variance supervision: when labels are scarce, learning a high-capacity classifier boundary is risky because the decision boundary can be dominated by a few labeled points, yielding unstable generalization. Prototypes provide a lower-variance alternative because they average labeled embeddings within each class, and even with limited data, the prototype can still point in a roughly correct direction in embedding space. Second, prototype alignment encourages embeddings to tilt toward their class centers, which for a similarity-based retrieval system primarily changes the \emph{top} of the ranked list: it increases the chance that the single closest neighbor is label-consistent, directly improving Hit@1 and MRR@K.

Third, the contrastive component prevents ``prototype collapse.'' If we only align to the correct prototype, the model could increase similarity to all prototypes by inflating common components, which would not improve discrimination. Contrast explicitly penalizes similarity to other prototypes, so conceptually, the model learns not only what the class looks like (alignment) but also what it is not (contrast), using prototypes as a compact set of negative anchors. Fourth, reconstruction and regularization preserve utility: a critical risk in geometry reshaping is collapse, where a transformation that improves one metric can harm others by destroying within-class variation. Reconstruction losses and orthogonality constraints mitigate this by preserving information about the original embedding while allocating class-discriminative signal into a controllable subspace. Finally, the metric differences observed in the results reflect different optimization targets: Purity@K measures neighborhood label-consistency, while Hit@K (for larger $K$) measures recall. PEARL prioritizes the very closest neighbors, explaining its strength in Hit@1 and MRR rather than Hit@20.

In practice, \method\ is most effective in label-scarce environments and for tasks that rely on cosine similarity neighborhoods (retrieval, ranking, kNN-style classifiers). It is particularly useful when labeling arrives incrementally, because prototypes provide a stable supervision signal and PEARL can be updated without a full encoder fine-tuning cycle. It is also a good fit for human-in-the-loop workflows where users inspect only the first few retrieved items, making Hit@1 and MRR@K operationally meaningful. Prototype alignment works best when classes have coherent ``centers'' in embedding space; if classes are highly multi-modal, PEARL may require multiple prototypes or a more expressive downstream model.

When labels are abundant, supervised projections such as LDA+L2 can dominate because they leverage richer class statistics to maximize separation. These observations highlight the value of reporting multiple metrics. \method\ can serve as a \emph{label-efficient warm start} for later, more supervised stages: in an evolving deployment, a team might start with a few hundred labels and apply PEARL, then later transition to stronger supervised projections as more labels become available. PEARL is also complementary to contrastive fine-tuning approaches---contrastive training can improve sentence embeddings when paired data or augmentation is available, but requires more training infrastructure. PEARL provides a lightweight alternative when only class labels are available.

\section{Industry and System Implications}
\label{sec:implications}

A typical deployment loop collects a small labeled set for the target taxonomy (even a few hundred labeled messages can be useful), computes prototypes from labeled embeddings and trains the PEARL refinement mapping, batch-transforms embeddings for the historical corpus and builds (or rebuilds) a retrieval index, and then monitors early retrieval quality using a small labeled validation stream (Hit@1/MRR) to trigger updates when drift is detected. Because PEARL operates on embeddings rather than raw text, it can be integrated into pipelines where text cannot be stored (for privacy reasons) but embeddings can be cached. The refinement model itself is small and can be executed efficiently on commodity hardware, allowing systems to update retrieval behavior without redeploying the base encoder.

In digital governance workflows, improved neighborhood quality translates into better routing and matching of citizen messages. Better early retrieval means fewer irrelevant cases shown to operators, reducing time spent on manual search; prototype-based alignment naturally supports auditing, as prototypes can be logged and compared over time to detect drift; and when a new complaint trend appears, a small amount of targeted labeling can reshape the retrieval space. Governance systems also benefit from interpretability---when a recommendation is produced by retrieving similar cases, it is easier to explain than a purely parametric prediction. PEARL strengthens this by making retrieved neighbors more label-consistent, reducing the risk of misleading evidence in high-stakes public-sector settings \citep{oneil2016weapons, eubanks2018automating, barocas2019fairness}.

The same pipeline applies to customer support, enterprise document retrieval, and compliance monitoring. In customer support, PEARL can improve ``similar ticket'' retrieval, enabling faster resolution. In compliance monitoring, where new risk categories appear episodically, a label-efficient neighborhood correction can reduce false alarms. Across domains, PEARL supports a common design pattern: keep the base encoder stable and improve downstream behavior through a lightweight transformation retrained with modest supervision.

\section{Limitations}
\label{sec:limitations}

\method\ requires at least some labeled data to form informative prototypes and guide alignment. It is not intended to outperform fully supervised projections in high-label regimes, and its behavior is tied to the quality and stability of prototypes under distribution shift. First, PEARL depends on prototype quality: if labeled data is extremely noisy, heavily imbalanced, or systematically biased, the resulting prototypes may be poor anchors, and PEARL may reshape the space in an undesirable direction. Prototype drift can also occur when the meaning of a category changes over time (e.g., a policy category expands to include new topics). Second, a single prototype per class is a coarse summary---classes that are inherently multi-modal (e.g., ``miscellaneous'' categories) may not be well represented by a single centroid. While PEARL can be extended to multiple prototypes per class, this introduces additional hyperparameters.

Third, there are metric tradeoffs: improving early precision (Hit@1/MRR) does not guarantee improving recall-style metrics at larger $K$. Systems that require high recall at large $K$ may prefer whitening-style baselines or hybrid pipelines. Finally, there is no guarantee under strong distribution shift. When the input distribution shifts substantially, prototypes computed from past labeled data may no longer represent current data. In such cases, PEARL should be paired with monitoring and periodic updates, potentially using active learning to refresh prototypes in newly emerging regions of the embedding space.

\section{Future Work}
\label{sec:future}

Future directions include extension to multi-label settings, nonlinear or adaptive prototype alignment, integration with contrastive fine-tuning \citep{gao2021simcse, yan2021consert}, and dynamic prototype updating in streaming systems. One direction is multi-label and hierarchical taxonomies: many real-world governance categories are multi-label (a message can involve multiple agencies or policy areas) and hierarchical (coarse-to-fine routing), and extending PEARL to this setting could involve learning multiple prototypes per instance, combining them with attention or mixture weights, and shaping neighborhoods to respect hierarchical relations. A second direction is adaptive and instance-conditioned alignment---prototype alignment need not be uniform, and a promising approach is to learn instance-conditioned alignment strength where examples that are already near their prototypes require minimal adjustment while ambiguous examples may require stronger correction. This could be implemented through gating functions or uncertainty-aware losses.

A third direction is dynamic prototypes and continual learning. In streaming systems, prototypes should evolve as language and topics shift, and a practical approach is to maintain prototypes with exponential moving averages, update them using newly labeled data, and detect drift when the distribution of similarities changes. This connects PEARL to continual learning and monitoring pipelines that are essential for governance deployments. Finally, future work should evaluate PEARL across additional domains and languages, and develop clearer theoretical characterizations of when prototype-guided alignment improves neighborhood structure---for example, analyzing how PEARL changes angular margins between classes could yield guidance for selecting temperatures and loss weights.

\section{Conclusion}
\label{sec:conclusion}

\method\ fills a gap between unsupervised embedding post-processing and fully supervised projection methods. By using limited labels to reshape representation geometry around class prototypes, it improves neighborhood quality and early retrieval precision in the label-scarce regimes that characterize many real-world systems---including digital governance communication platforms. The results underscore a broader lesson for embedding-based systems: improving representation geometry is not a cosmetic post-processing step, but a central design choice that can determine whether retrieval and case-based decision support behave reliably. Different post-processing methods optimize different parts of the ranked list---PEARL emphasizes label-consistent nearest neighbors and early ranking, while whitening and supervised projections can be preferable under different label regimes or retrieval objectives. Overall, PEARL provides a practical and label-efficient approach for improving fixed embeddings without retraining the base encoder, making it well-suited for deployment settings where labeling is expensive, the data distribution evolves, and systems must be improved incrementally with minimal disruption.

\bibliography{references}

\end{document}